\documentclass[preprint,authoryear,12pt]{elsarticle}

\usepackage{multicol}
\usepackage{multirow}
\usepackage{graphicx}
\usepackage{algorithmic}
\usepackage{algorithm}
\usepackage{amsmath}
\usepackage{amssymb}

\begin{document}

\begin{frontmatter}

\title{Term-Weighting Learning via Genetic Programming for Text Classification}
\author[inaoe]{Hugo Jair Escalante\corref{cor1}}\ead{hugojair@inaoep.mx}
\author[inaoe]{Mauricio A. Garc\'ia-Lim\'on}\ead{ 	mauricio.garcia.cs@gmail.com }
\author[inaoe]{Alicia Morales-Reyes}\ead{a.morales@inaoep.mx}
\author[umsnh]{Mario Graff}\ead{mgraffg@gmail.com}
\author[inaoe]{Manuel Montes-y-G\'omez}\ead{mmontesg@inaoep.mx}
\author[inaoe]{Eduardo F. Morales}\ead{emorales@inaoep.mx}
\address[inaoe]{Computer Science Department,\\ Instituto Nacional de Astrof\'isica, \'Optica y Electr\'onica,\\
Luis Enrique Erro 1, Puebla, 72840, Mexico}
\address[umsnh]{Divisi\'on de Estudios de Postgrado \\Facultad de
Ingenier\'ia El\'ectrica \\Universidad Michoacana de San Nicol\'as de
Hidalgo, M\'exico}
\cortext[cor1]{Corresponding author.}
\begin{abstract}
This paper describes a novel approach to learning term-weighting schemes (TWSs) in the context of text classification. In text mining a TWS determines the way in which documents will be represented in a vector space model, before applying a classifier.  Whereas acceptable performance has been obtained with standard TWSs (e.g., Boolean and term-frequency schemes), the definition of TWSs has been traditionally an art. Further, it is still a difficult task to determine what is the best TWS for a particular problem and it is not clear yet, whether better schemes, than those currently available, can be generated by combining known TWS.
We propose in this article a genetic program that aims at learning effective TWSs that can improve the performance of current schemes in text classification. The genetic program learns how to combine a set of basic units to give rise to discriminative TWSs. We report an extensive experimental study comprising data sets from thematic and non-thematic text classification as well as from image classification. Our study shows the validity of the proposed method; in fact, we show that TWSs learned with the genetic program outperform traditional schemes and other TWSs proposed in recent works. Further, we show that TWSs learned from a specific domain can be effectively used for other tasks.
\end{abstract}
\end{frontmatter}
\section{Introduction}
\label{sec:intro}

Text classification (TC) is the task of associating documents with predefined categories that are related to their content. TC is an important and active research field because of the large number of digital documents available and the consequent need to organize them.
The TC problem has been approached with pattern classification methods, where documents are represented as numerical vectors and standard classifiers (e.g., na\"ive Bayes and support vector machines) are applied~\citep{sebastiani}. This type of representation is known as the vector space model (VSM)~\citep{Salton}. Under the VSM one assumes a document is a point in a  $N$-dimensional space and documents that are closer in that space are similar to each other~\citep{Turney}.  Among the different instances of the VSM, perhaps the most used model is the bag-of-words (BOW) representation.  In the BOW it is assumed that the content of a document can be determined by the (orderless) set of terms\footnote{A term is any basic unit by which documents are formed, for instance, terms could be words, phrases, and sequences (n-grams) of words or characters. } it contains.
Documents are represented as points in the vocabulary space, that is,  a document is represented by a numerical vector of length equal to the number of different terms in the vocabulary (the set of all different terms in the document collection). The elements of the vector specify how important the corresponding terms are for describing the semantics or the content of the document. BOW is the most used document representation in both TC and information retrieval. In fact, the BOW representation has been successfully adopted for processing other media besides text, including, images~\citep{csurkabow}, videos~\citep{videoRetrieval}, speech signals~\citep{audioBOW}, and time series~\citep{TimeSeriesBOW} among others.

A crucial component of TC systems using the BOW representation is the term-weighting scheme (TWS), which is in charge of determining how relevant a term is for describing the content of a document~\citep{textmining,analytical,PAMI1,Super}. Traditional TWS are term-frequency (\emph{TF}), where the importance of a term in a document is given by its frequency of occurrence in the document; Boolean (\emph{B}), where the importance of a term in document is either 1, when the term appear in the document or 0, when the term does not appear in the document; and term-frequency inverse-document-frequency (\emph{TF-IDF}), where the importance of a term for a document is determined by its occurrence frequency times the inverse frequency of the term across the corpus (i.e., frequent terms in the corpus, as prepositions and articles, receive a low weight).  Although, TC is a widely studied topic with very important developments in the last two decades~\citep{sebastiani,textmining}, it is somewhat surprising that little attention has been paid to the development of new TWSs to better represent the content of documents for TC. In fact, it is quite common in TC systems that researchers use one or two common TWSs (e.g., \emph{B}, \emph{TF} or \emph{TF-IDF}) and put more effort in other processes, like feature selection~\citep{George,featsel2}, or the learning process itself~\citep{TXMLSurvey,TCSurvey,psmsav}. Although all of the TC phases are equally important, we think that by putting more emphasis on defining or learning effective TWSs we can achieve substantial improvements in TC performance.

This paper introduces a novel approach to learning TWS for TC tasks. A genetic program is proposed in which a set of primitives and basic TWSs are combined  through arithmetic operators in order to generate alternative schemes that can improve the performance of a classifier. Genetic programming is a type of evolutionary algorithm in which a population of programs is evolved~\citep{gpbook}, where programs encode solutions to complex problems (mostly modeling problems), in this work programs encode TWSs.
The underlying hypothesis of our proposed method is that an evolutionary algorithm can learn TWSs of comparable or even better performance than those proposed so far in the literature.

Traditional TWSs combine term-importance and term-document-importance factors to generate TWSs. For instance in \emph{TF-IDF}, TF and IDF are term-document-importance and term-importance factors, respectively. Term-document weights are referred as local factors, because they account for the occurrence of a term in a document (locally). On the other hand, term-relevance weights are considered global factors, as they account for the importance of a term across the corpus (globally). It is noteworthy that
the actual factors that define a TWS and the combination strategy itself have been determined manually.
Herein we explore the suitability of learning these TWSs automatically, by providing a genetic program with a pool of TWSs' building blocks with the goal  of evolving a TWS that maximizes the classification performance for a TC classifier. We report experimental results in many TC collections that comprise both: thematic and non-thematic TC problems. Throughout extensive experimentation we show that the proposed approach is very competitive, learning very effective TWSs that outperform most of the schemes proposed so far. We evaluate the performance of the proposed approach under different settings and analyze the characteristics of the learned TWSs. Additionally, we evaluate the generalization capabilities of the learned TWSs and even show that a TWS learned from text can be used to effectively represent images under the BOW formulation.

The rest of this document is organized as follows. Next section formally introduces the TC task and describes common TWSs. Section~\ref{sec:rw} reviews related work on TWSs. Section~\ref{sec:method} introduces the proposed method. Section~\ref{sec:experiments} describes the experimental settings adopted in this work and reports results of experiments that aim at evaluating different aspects of the proposed approach. Section~\ref{sec:conclusions} presents the conclusions derived from this paper and outlines future research directions.

\section{Text classification with the Bag of Words}

The most studied TC problem is the so called thematic TC (or simply text categorization)~\citep{sebastiani}, which means that classes are associated to different themes or topics (e.g., classifying news into \emph{``Sports''} vs. \emph{``Politics''} categories). In this problem, the sole occurrence of certain terms may be enough to determine the topic of a document; for example, the occurrence of words/terms \emph{``Basketball'', ``Goal'', ``Ball'',} and \emph{``Football''} in a document is strong evidence that the document is about \emph{``Sports''}. Of course, there are more complex scenarios for thematic TC, for example, distinguishing documents about sports news into the categories: \emph{``Soccer''} vs. \emph{``NFL''}.
Non-thematic TC, on the other hand, deals with the problem of associating documents with labels that are not (completely) related to their topics. Non-thematic TC includes the problems of authorship attribution~\citep{stamatatos}, opinion mining and sentiment analysis~\citep{bopang}, authorship verification~\citep{authorver}, author profiling~\citep{profiling}, among several others~\citep{irony,doubleenten}.  In all of these problems, the thematic content is of no interest, nevertheless, it is common to adopt standard TWSs for representing documents in non-thematic TC as well (e.g., BOW using character n-grams or part-of-speech tags~\citep{stamatatos}).

It is noteworthy that the BOW representation has even trespassed the boundaries of the text media. Nowadays, images~\citep{csurkabow}, videos~\citep{videoRetrieval}, audio~\citep{audioBOW},  and other types of data~\citep{TimeSeriesBOW} are represented throughout analogies to the BOW. In non-textual data, a codebook is first defined/learned and then the straight BOW formulation is adopted.
In image classification, for example, visual descriptors extracted from images are clustered and the centers of the clusters are considered as visual words~\citep{csurkabow,BOWImages}. Images are then represented by numerical vectors (i.e., a VSM) that indicate the relevance of visual words for representing the images. Interestingly, in other media than text (e.g., video, images)
it is standard to use only the \emph{TF} TWS, hence motivating the study on the effectiveness of alternative TWSs in non-textual tasks. Accordingly, in this work we also perform experiments on learning TWSs for a standard computer vision problem~\citep{caltech}.

TC is a problem that has been approached mostly as a supervised learning task, where the goal is to learn a model capable of associating documents to categories~\citep{sebastiani,textmining,TXMLSurvey}.
Consider a data set of labeled documents $\mathcal{D} = (\textbf{x}_i, y_i)_{\{1, \ldots, N\}}$ with $N$ pairs of documents ($\textbf{x}_i$)  and their classes ($y_i$) associated to a TC problem; where we assume $\textbf{x}_i\in \mathbb{R}^p$ (i.e., a VSM) and $y_i \in C = \{1, \ldots K\}$, for a problem with $K-$classes. The goal of TC is to learn a function $f : \mathbb{R}^{p} \rightarrow C$ from $\mathcal{D}$ that can be used to make predictions for documents with unknown labels, the so called test set: $\mathcal{T} = \{\textbf{x}^T_1, \ldots, \textbf{x}^T_M\}$.
Under the BOW formulation, the dimensionality of documents' representation, $p$, is defined as $p=|V|$, where $V$ is the vocabulary (i.e., the set all the different terms/words that appear in a corpus). Hence,
each document $d_i$ is represented by a numerical vector $\mathbf{x}_i = \langle x_{i,1} \ldots, x_{i,|V|} \rangle$, where each element  $x_{i,j}$, $j=1,\ldots,|V|$, of $\mathbf{x}_i$ indicates how relevant word $t_j$ is for describing the content of $d_i$, and where the value of $x_{i,j}$ is determined by the TWS.

Many TWSs have been proposed so far, including unsupervised~\citep{sebastiani,Salton,textmining} and supervised schemes~\citep{Super,PAMI1}, see Section~\ref{sec:rw}. Unsupervised TWSs are the most used ones, they were firstly proposed for information retrieval tasks and latter adopted for TC~\citep{sebastiani,Salton}.  Unsupervised schemes rely on term frequency statistics and measurements that do not take into account any label information. For instance, under the Boolean (B) scheme $x_{i,j}=1$ $iff$ term $t_j$ appears in document $i$ and 0 otherwise; while in the term-frequency (\emph{TF}) scheme, $x_{i,j} = \#(d_i,t_j)$, where $\#(d_i,t_j)$ accounts for the times term $t_j$  appears in document $d_i$. On the other hand, supervised TWSs aim at incorporating discriminative information into the representation of documents~\citep{Super}. For example in the \emph{TF-IG} scheme, $x_{i,j} = \#(d_i,t_j) \times IG(t_j)$, is the product of the TF TWS for term $t_j$ and document $d_i$ (a local factor) with the information gain of term $t_j$ ($IG(t_j)$, global factor). In this way, the discrimination power of each term is taken into account for the document representation; in this case through the information gain value~\citep{featsel2}. It is important to emphasize that most TWSs combine information from both term-importance (global) and term-document-importance (local) factors (see Section~\ref{sec:rw}), for instance, in the \emph{TF-IG} scheme, IG is a term-importance factor, whereas TF is a term-document-importance factor.

Although acceptable performance has been reported with existing TWS, it is still an art determining the adequate TWS for a particular data set; as a result, mostly unsupervised TWSs (e.g., \emph{B, TF} and  \emph{TF-IDF}) have been adopted for TC systems~\citep{textmining,TCSurvey}. A first hypothesis of this work is that different TWSs can achieve better performance on different TC tasks (e.g., thematic TC vs. non-thematic TC); in fact, we claim that within a same domain (e.g., news classification) different TWSs are required to obtain better classification performance on different data sets. On the other hand, we notice that TWSs have been defined as combinations of term-document weighting factors (which can be seen as other TWSs, e.g., \emph{TF}) and  term-relevance measurements (e.g., \emph{IDF} or \emph{IG}), where the definition of TWSs has been done by relying on the expertise of users/researchers. Our second hypothesis is that 
the definition of new TWSs can be automated. 
With the aim of verifying both hypotheses, this paper introduces a genetic program that learns how to combine term-document-importance and term-relevance factors to generate effective TWSs for diverse TC tasks.

\section{Related work}
\label{sec:rw}

As previously mentioned, in TC it is rather common to use unsupervised TWSs to represent documents, specifically \emph{B, TF} and \emph{TF-IDF} schemes are very popular (see Table~\ref{tab:pesados}). Their popularity derives from the fact that these schemes have proved to be very effective in information retrieval~\citep{Salton,ModernIR,Turney} and in many TC problems as well as~\citep{sebastiani,textmining,TXMLSurvey,TCSurvey,miningtd}. Unsupervised TWSs mainly capture term-document occurrence (e.g., term occurrence frequency, \emph{TF}) and term-relevance (e.g., inverse document frequency, \emph{IDF}) information. While acceptable performance has been obtained with such TWSs in many applications, in TC one has available labeled documents, and hence, document-label information can also be exploited to obtain more discriminative TWSs. This observation was noticed by Debole \& Sebastiani and other authors that have introduced supervised TWSs~\citep{Super,PAMI1}.

Supervised TWSs take advantage of labeled data by incorporating a discriminative term-weighting factor into the TWSs. In~\citep{Super} TWSs were defined by combining the unsupervised TF scheme with the following term-relevance criteria: information gain (\emph{TF-IG}), which measures the reduction of entropy when using a term as classifier~\citep{featsel2}; $\chi^2$ (\emph{TF-CHI}), makes an independence test regarding a term and the classes~\citep{sebastiani}; and gain-ratio (\emph{TF-GR}) measuring the gain-ratio when using the term as classifier~\citep{Super}. The conclusions from~\citep{Super} were that small improvements can be obtained with supervised TWSs over unsupervised ones. Although somewhat disappointing, it is interesting that for some scenarios supervised TWSs were beneficial. More recently, Lan et al. proposed an alternative supervised TWS~\citep{PAMI1}, the so called \emph{TF-RF} scheme. \emph{TF-RF} combines \emph{TF} with a criterion that takes into account the true positives and true negative rates when using the occurrence of the term as classifier. In~\citep{PAMI1} the proposed \emph{TF-RF} scheme obtained better performance than unsupervised TWSs and even outperformed the schemes proposed in~\citep{Super}. In~\citep{analytical} the \emph{RF} term-relevance factor was compared with alternative weights, including mutual information, odds ratio and $\chi^2$; in that work\emph{RF} outperformed the other term-importance criteria.

Table~\ref{tab:pesados} shows most of the TWSs proposed so far for TC.
It can be observed that TWSs are formed by combining term-document (\emph{TDR}) and term (\emph{TR}) relevance weights. The selection of what TDR and TR weights to use rely on researchers choices (and hence on their biases). It is quite common to use \emph{TF} as \emph{TDR}, because undoubtedly the term-occurrence frequency carries on very important information: we need a way to know what terms a document is associated with. However, it is not that clear what \emph{TR} weight to use, as there is a wide variety of TR factors that have been proposed. The goal of \emph{TRs} is to determine the importance of a given term, with respect to the documents in a corpus (in the unsupervised case) or to the classes of the problem (in the supervised case). Unsupervised \emph{TRs} include: global term-frequency, and inverse document frequency (IDF) TRs. These weights can capture word importance depending on its global usage across a corpus, however, for TC seems more appealing to use discriminative TRs as one can take advantage of training labeled data. In this aspect, there is a wide variety of supervised TRs that have been proposed, including: mutual information, information gain, odds ratio, etcetera~\citep{miningtd}.
\begin{table}[htb]
\centering
\caption{Common term weighting schemes for TC. In every TWS, $x_{i,j}$ indicates how relevant term $t_j$ is for describing the content of document $d_i$, under the corresponding TWS. $N$ is the number of documents in training data set, $\#(d_i,t_j)$ indicates the frequency of term $t_j$ in document $d_i$, $df(t_j)$ is the number of documents in which term $t_j$ occurs, $IG(t_j)$ is the information gain of term $t_j$, $CHI(t_j)$ is the $\chi^2$ statistic for term $t_j$, and $TP$, $TN$  are the true positive and true negative rates for term $t_j$ (i.e., the number of positive, resp. negative, documents that contain term $t_j$).
}\label{tab:pesados}
\tiny{
\begin{tabular}{|p{0.8cm}|p{1.5cm}|p{2.5cm}|p{3cm}|p{2cm}|}
\hline
\textbf{Acronym}&\textbf{Name}&\textbf{Formula}&\textbf{Description}&\textbf{Ref.}\\\hline
\emph{B}& Boolean& $x_{i,j} = \mathbf{1}_{\{\#(t_i,d_j)>0\}}$
&Indicates the prescense/abscense of terms&\citep{Salton}\\
\emph{TF}& Term-Frequency& $x_{i,j} = \#(t_i,d_j)$&Accounts for the frequency of occurrence of terms&\citep{Salton}\\
\emph{TF-IDF}& TF - Inverse Document Frequency& $x_{i,j} = \#(t_i,d_j) \times \log(\frac{N}{df(t_j)})$ &An TF scheme that penalizes the frequency of terms across the collection&\citep{Salton}\\
\emph{TF-IG}& TF - Information Gain& $x_{i,j} = \#(t_i,d_j) \times IG(t_j)$
&TF scheme that weights term occurrence by its information gain across the corpus. &\citep{Super}\\
\emph{TF-CHI}& TF - Chi-square& $x_{i,j} = \#(t_i,d_j) \times CHI(t_j)$ &TF scheme that weights term occurrence by its $\chi^2$ statistic &\citep{Super}\\
\emph{TF-RF}& TF - Relevance Frequency& $x_{i,j} = \#(t_i,d_j) \times \log( 2 + \frac{TP}{\max (1,TN)})$ &TF scheme that weights term occurrence by its $\chi^2$ statistic &\citep{PAMI1}\\\hline
\end{tabular}}
\end{table}

The goal of a supervised TR weight is to determine the importance of a given term with respect to the classes. The simplest, TR would be to estimate the correlation of term frequencies and the classes, although any other criterion that accounts for the association of terms and classes can be helpful as well.
It is interesting that although many TRs are available out there, they have been mostly used for feature selection rather than for building TWSs for TC.  Comprehensive and extensive comparative studies using supervised TRs for feature selection have been reported~\citep{analytical,George,featsel2,featsel}. Although not being conclusive, these studies serve to identify the most effective TRs weights, such weights are considered in this study.

To the best of our knowledge, the way we approach the problem of learning TWSs for TC is novel. Similar approaches based on genetic programming to learn TWSs have been proposed in~\citep{GPIR,GPTX1,GPTX2,trotman,oren,fana}, however, these researchers have focused on the information retrieval problem, which differs significantly from TC. Early approaches using genetic programming to improve the \emph{TF-IDF} scheme for information retrieval include those from~\citep{trotman,oren,fana,fanb}. More recently, Cummins et al. proposed improved genetic programs to learn TWSs also for information retrieval~\citep{GPIR,GPTX1,GPTX2}.

Although the work by Cummins et al. is very related to ours, there are major differences (besides the problem being approached):
Cummins et al. approached the information retrieval task and defined a TWS as a combination of three factors: local, global weighting schemes and a normalization factor\footnote{Recall a local factor incorporates term information (locally) available in a document, whereas a global term factor takes into account term statistics estimated across the whole corpus. In information retrieval it is also common to normalize the vectors representing a document to reduce the impact of the length of a document.}.
The authors designed a genetic program that aimed at learning a TWS by evolving the local and global schemes separately. Only 11 terminals, including constants, were considered. Since information retrieval is an unsupervised task, the authors have to use a whole corpus with relevance judgements (i.e., a collection of documents with queries and the set of relevant documents to each query) to learn the TWS, which, once learned, could be used for other information retrieval tasks. Hence they require a whole collection of documents to learn a TWS. On the other hand, the authors learned a TWS separately, first a global TWS was evolved fixing a binary local scheme, then a local scheme was learned by fixing the learned global weight. Hence, they restrict the search space for the genetic program, which may limit the TWSs that can be obtained. Also, it is worth noticing that the focus of the authors of~\citep{GPIR,GPTX1,GPTX2} was on learning a single, and generic TWS to be used for other information retrieval problems, hence the authors performed many experiments and reported the single best solution they found after extensive experimentation. Herein, we provide an extensive evaluation of the proposed approach, reporting average performance over many runs and many data sets. Finally, one should note that the approach from~\citep{GPIR,GPTX1,GPTX2} required of large populations and numbers of generations (1000 individuals and 500 generations were used), whereas in this work competitive performance is obtained with only 50 individuals and 50 generations.

\section{Learning term-weighting schemes via GP}
\label{sec:method}

As previously mentioned, the traditional approach for defining TWSs has been somewhat successful so far. Nevertheless, it is still unknown whether we can automatize the TWS definition process and obtain TWSs of better classification performance in TC tasks.
In this context, we propose a genetic programming solution that aims at learning effective TWSs automatically.
We provide the genetic program with a pool of TDR and TR weights as well as other TWSs and let a program search for the TWS that maximizes an estimate of classification performance. Thus, instead of defining TWSs based on our own experiences on text mining, we let a computer itself to build an effective TWS. The advantages of this approach are that it may allow to learn a specific TWS for each TC problem, or to learn TWSs from one data set (e.g., a small one) and implement it in a different collection (e.g., a huge one). Furthermore, the method reduces the dependency on users/data-analysts and their degree of expertise and biases for defining TWSs. The rest of this section describes the proposed approach.
We start by providing a brief overview of genetic programming, then we explain in detail the proposal, finally, we close this section with a discussion on the benefits and limitations of our approach.

\subsection{Genetic programming}

\begin{figure}[htb!]
    \centering
    \includegraphics[scale=0.6]{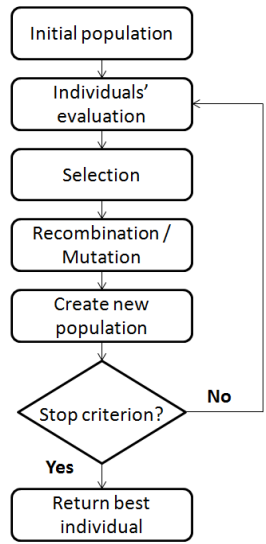}
    \caption{A generic diagram of an evolutionary algorithm. }
  \label{fig:genEA}
\end{figure}

Genetic programming (GP)~\citep{gpbook} is an evolutionary technique which follows the reproductive cycle of other evolutionary algorithms such as genetic algorithms (see Figure~\ref{fig:genEA}): an initial population is created (randomly or by a pre-defined criterion), after that, individuals are selected, recombined, mutated and then placed back into the solutions pool.
The distinctive feature of GP, when compared to other evolutionary algorithms, is in that complex data structures are used to represent solutions (individuals), for example, trees or graphs. As a result, GP can be used for solving complex learning/modeling problems.
In the following we describe the GP approach to learn TWSs for TC.

\subsection{TWS learning with genetic programming}

We face the problem of learning TWSs as an optimization one, in which we want to find a TWSs that maximizes the classification performance of a classifier trained with the TWS. We define a valid TWS as the combination of: (1) other TWSs, (2) TR and (3) TDR factors, and restrict the way in which such components can be combined by a set of arithmetic operators. We use GP as optimization strategy, where each individual corresponds to a tree-encoded TWS.  The proposed genetic program
explores the search space of TWSs that can be generated by combining TWSs, TRs and TDRs with a predefined set of operators. The rest of this section details the components of the proposed genetic program, namely, representation, terminals and function set, genetic operators and fitness function.

\subsubsection{Representation}
\begin{figure}[htb!]
    \centering
    \includegraphics[scale=0.45]{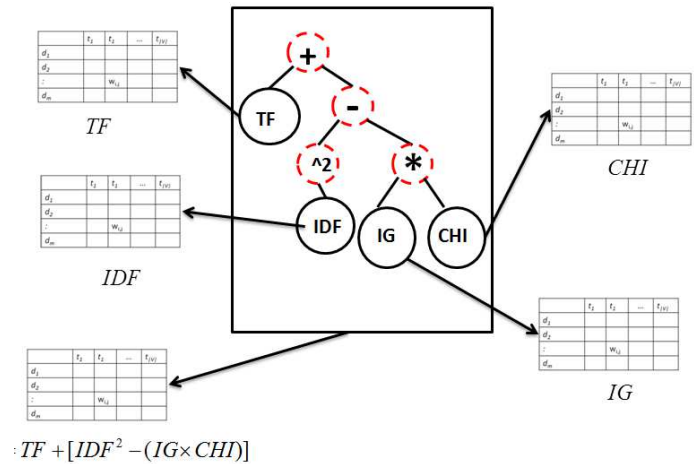}
    \caption{Representation adopted for TWS learning. }
  \label{fig:representation}
\end{figure}

Solutions to our problem are encoded as trees, where we define terminal nodes to be the building blocks of TWSs. On the other hand, we let internal nodes of trees to be instantiated by arithmetic operators that combine the building blocks to generate new TWSs. The representation is graphically described in Figure~\ref{fig:representation}.

\subsubsection{Terminals and function set}
As previously mentioned, traditional TWSs are usually formed by two factors: a term-document relevance (\emph{TDR}) weight and a term-relevance (\emph{TR}) factor.  The most used \emph{TDR} is term frequency (\emph{TF}), as allows one to relate documents with the vocabulary. We consider \emph{TF} as \emph{TDR} indicator, but also we consider standard TWSs (e.g., \emph{Boolean, TD, RF}) as \emph{TDR} weights. The decision to include other TWSs as building blocks is in order to determine whether standard TWSs can be enhanced with GP. 
Regarding \emph{TR}, there are many
alternatives available.  In this work we analyzed the most common and effective \emph{TR} weights as reported in the literature~\citep{sebastiani,analytical,PAMI1,Super,George} and considered them as building blocks for generating TWSs.
Finally we also considered some constants as building blocks. The full set of building blocks (terminals in the tree representation) considered is shown in Table~\ref{tab:pesados}, whereas the set of operators considered in the proposed method (i.e., the function set) is the following:  $\mathcal{F}= \{ +, -, *, /, \log_2{x}, \sqrt{x}, x^2\}$, where $\mathcal{F}$ includes operators of arities one and two.
\begin{table}[!h]
  \centering
  \caption {Terminal set.}\label{tabla:params}
  \begin{center}
  \tiny{
                  	\begin{tabular}[t]{| c | c |}
                		\hline
                    		\textbf{Variable} & \textbf{Meaning} \\
                		\hline
							$\mathbf{W}_1$ & $N$, Constant matrix, the total number of training documents. \\
                		\hline
							$\mathbf{W}_2$ & $\|V\|$, Constant matrix, the number of terms. \\
                		\hline
							$\mathbf{W}_3$ & $CHI$, Matrix containing in each row the vector of $\chi^2$ weights for the terms.\\
                		\hline
							$\mathbf{W}_4$ & $IG$, Matrix containing in each row the vector of information gain weights for the  terms.\\
                		\hline
							$\mathbf{W}_5$ & $TF-IDF$, Matrix with the TF-IDF term weighting scheme. \\
                		\hline
							$\mathbf{W}_6$ & $TF$, Matrix containing the TF term-weighting scheme. \\
                		\hline
							$\mathbf{W}_7$ & $FGT$, Matrix containing in each row the global term-frequency for all terms. \\
                		\hline
							$\mathbf{W}_8$ & $TP$, Matrix containing in each row the vector of true positives for all terms.  \\
                		\hline
							$\mathbf{W}_9$ & $FP$, Matrix containing in each row the vector of false positives.\\
                		\hline
							$\mathbf{W}_{10}$ & $TN$, Matrix containing in each row the vector of true negatives.\\
                		\hline
							$\mathbf{W}_{11}$ & $FN$, Matrix containing in each row the vector of false negatives.\\
            			\hline
            			    $\mathbf{W}_{12}$ & $Accuracy$, Matrix in which each row contains the accuracy obtained when using the term as classifier.\\
            			\hline
							$\mathbf{W}_{13}$ & $Accuracy\_Balance$, Matrix containing the AC\_Balance each (term,  class).\\
            			\hline
							$\mathbf{W}_{14}$ & $BNS$, An array that contains the value for each BNS per (term, class).\\
						\hline
							$\mathbf{W}_{15}$ & $DFreq$, Document frequency matrix containing the value for each (term, class).\\
						\hline
							$\mathbf{W}_{16}$ & $FMeasure$, F-Measure matrix containing the value for each (term, class).\\
						\hline
							$\mathbf{W}_{17}$ & $OddsRatio$, An array containing the OddsRatio term-weighting.\\
						\hline
							$\mathbf{W}_{18}$ & $Power$, Matrix containing the Power value for each (term, class).\\
						\hline
							$\mathbf{W}_{19}$ & $ProbabilityRatio$, Matrix containing the ProbabilityRatio each (term, class).\\
						\hline
							$\mathbf{W}_{20}$ & $Max\_Term$, Matrix containing the vector with the highest repetition for each term.\\
						\hline
							$\mathbf{W}_{21}$ & $RF$, Matrix containing the RF vector. \\
						\hline
							$\mathbf{W}_{22}$ & $TF \times RF$, Matrix containing $TF \times RF$. \\
						\hline
          \end{tabular}}
  \end{center}
\end{table}		

In the proposed approach, a TWS is seen as a combination of building blocks by means of arithmetic operators. One should note, however, that three types of building blocks are considered: \emph{TDR}, \emph{TR} and constants. Hence we must define a way to combine matrices (\emph{TDR} weights), vectors (\emph{TR} scores) and scalars (the constants), in such a way that the combination leads to a TWS (i.e., a form of \emph{TDR}). Accordingly, and for easiness of implementation, each building block shown in Table~\ref{tab:pesados} is processed as a matrix of the same length as the TWS (i.e., $N \times |V|$) and operations are performed element-wise. In this way a tree can be directly evaluated, and the operators are applied between each element of the matrices, leading to a TWS.

\emph{TDRs} are already matrices of the same size as the TWSs: $N \times |V|$. In the case of \emph{TRs}, we have a vector of length $|V|$, thus for each \emph{TR} we generate a matrix of size $N \times |V|$ where each of its rows is the \emph{TR}; that is, we repeat $N$ times the TR weight. In this way, for example, a TWS like \emph{TF-IDF} can be obtained as $TF \times IDF$, where the $\times$ operator means that each element $tf_{i,j}$ of \emph{TF} is multiplied by each element of the \emph{IDF} matrix $idf_{i,j}$ and where $idf_{i,j} = \log(\frac{N}{df(t_j)})$ for $i = 1, \ldots, N$, all \emph{TRs} were treated similarly. In the case of constants we use a scalar-matrix operator, which means that the constant is operated with each element of the matrix under analysis.

Estimating the matrices each time a tree is evaluated can be a time consuming process, therefore, at the beginning of the search process we compute the necessary matrices for every terminal from Table~\ref{tab:pesados}. Hence, when evaluating an individual we only have to use the values of the precomputed matrices and apply the operators specified by a tree.

\subsubsection{Genetic operators}
As explained above, in GP a population of individuals is initialized and evolved according to some operators that aim at improving the quality of the population. For initialization we used the standard ramped-half-and-half strategy~\citep{Eiben}, which generates half of the population with (balanced) trees of maximum depth, and the other half with trees of variable depth. As genetic operators we also used standard mechanisms: we considered the subtree crossover and point mutation. The role of crossover is to take two promising solutions and combine their information to give rise to two offspring, with the goal that the offspring have better performance than the parents. The subtree crossover works by selecting two parent solutions/trees (in our case, via tournament) and randomly select an internal node in each of the parent trees. Two offspring are created by interchanging the subtrees below the identified nodes in the parent solutions.

The function of the mutation operator is to produce random variations in the population, facilitating the exploration capabilities of GP. The considered mutation operator first selects an individual to be mutated. Next an internal node of the individual is identified, and if the internal node is an operator (i.e., a member of $\mathcal{F}$) it is replaced by another operator of the same arity. If the chosen node is a terminal, it is replaced by another terminal. Where in both cases the replacing node is selected with uniform probability.

\subsubsection{Fitness function}

As previously mentioned, the aim of the proposed GP approach is to generate a TWS that obtains competitive classification performance. In this direction, the goodness of an individual is assessed via the classification performance of a predictive model that uses the representation generated by the TWS. Specifically, given a solution to the problem,
we first evaluate the tree to generate a TWS using the training set. Once training documents are represented by the corresponding TWS, we perform a $k-$fold cross-validation procedure
to assess the effectiveness of the solution. In $k-$fold cross validation, the training set is split into $k$ disjoint subsets, and $k$ rounds of training and testing are performed; in each round $k-1$ subsets are used as training set and 1 subset is used for testing, the process is repeated $k$ times using a different subset for testing each time. The average classification performance is directly used as fitness function. Specifically, we evaluate the performance of classification models with the $f_1$ measure. Let $TP$, $FP$ and $FN$ to denote the true positives, false positives and false negative rates for a particular class, precision ($Prec$) is defined as $\frac{TP}{TP +FP}$ and recall ($Rec$) as $\frac{TP}{TP + FN}$. $f_1$-measure is simply the harmonic average between precision and recall: $f_1 = \frac{2 \times Prec \times Rec}{Prec + Rec}$. The average across classes is reported (also called, macro-average $f_1$), this way of estimating the $f_1$-measure is known to be particularly useful when tackling unbalanced data sets~\citep{sebastiani}.

Since under the fitness function $k$ models have to be trained and tested for the evaluation of a single TWS, we need to look for an efficient classification model that, additionally, can deal naturally with the high-dimensionality of data. Support vector machines (SVM) comprise a type of models that have proved to be very effective for TC~\citep{sebastiani,TCsvm}. SVMs can deal naturally with the sparseness and high dimensionality of data,  however, training and testing an SVM can be a time consuming process. Therefore, we opted for efficient implementations of SVMs that have been proposed recently~\citep{LLSVM,budgetSVM}. That methods are trained online and under the scheme of learning with a budget. We use the  predictions of an SVM as the fitness function for learning TWSs. Among the methods available in~\citep{budgetSVM} we used the low-rank linearized SVM (LLSMV)~\citep{LLSVM}. LLSVM is a linearized version of non-linear SVMs, which can be trained efficiently with the so called block minimization framework~\citep{BMF}. We selected LLSVM instead of alterative methods, because this method has outperformed several other efficient implementations of SVMs, see e.g.,~\citep{budgetSVM,LLSVM}.

\subsection{Summary}

We have described the proposed approach to learn TWSs via GP. When facing a TC problem we start by estimating all of the terminals described in Table~\ref{tab:pesados} for the training set. The terminals are feed into the genetic program, together with the function set. We used the GPLAB toolbox for implementing the genetic program with default parameters~\citep{gplab}. The genetic program searches for the tree that maximizes the $k-$fold cross validation performance of an efficient SVM using training data only. After a fixed number of generations, the genetic program returns the best solution found so far, the best TWS. Training and test (which was not used during the search process) data sets are represented according to such TWS. One should note that all of the supervised term-weights in Table~\ref{tab:pesados} are estimated from the training set only (e.g., the information gain for terms is estimated using only the labeled training data); for representing test data we use the pre-computed term-weights. Next, the LLSVM is trained in training data  and the trained model makes predictions for test samples. We evaluate the performance of the proposed method by comparing the predictions of the model and the actual labels for test samples.  The next section reports results of experiments that aim at evaluating the validity of the proposed approach.

\section{Experiments and results}
\label{sec:experiments}

This section presents an empirical evaluation of the proposed TWL approach. The goal of the experimental study is to assess the effectiveness of the learned TWSs and compare their performance to existing schemes. Additionally, we evaluate the generalization performance of learned schemes, and their effectiveness under different settings.

\subsection{Experimental settings}

For experimentation we considered a suite of benchmark data sets associated to three types of tasks: thematic TC, authorship attribution (AA, a non-thematic TC task) and image classification (IC). Table~\ref{tab:datasets} shows the characteristics of the data sets. We considered three types of tasks because we wanted to assess the generality of the proposed approach.

\begin{table}[htb]
\centering
\caption{Data sets considered for experimentation}\label{tab:datasets}
\footnotesize{
\begin{tabular}{|c|c|c|c|c|}
\hline
\multicolumn{5}{|c|}{\textbf{Text categorization}}\\\hline
\textbf{Data set}&\textbf{Classes}&\textbf{Terms}&\textbf{Train}&\textbf{Test}\\\hline
Reuters-8&8&23583&5339&2333\\[-0.05in]
Reuters-10&10&25283&6287&2811\\[-0.05in]
20-Newsgroup&20&61188&11269&7505\\[-0.05in]
TDT-2&30&36771&6576&2818\\[-0.05in]
WebKB&4&7770&2458&1709\\[-0.05in]
Classic-4&4&5896&4257&2838\\[-0.05in]
CADE-12&12&193731&26360&14618\\\hline
\multicolumn{5}{|c|}{\textbf{Authorship attribution}}\\\hline
\textbf{Data set}&\textbf{Classes}&\textbf{Terms}&\textbf{Train}&\textbf{Test}\\\hline
CCA-10&10&15587&500&500\\[-0.05in]
Poetas&5&8970&71&28\\[-0.05in]
Football&3&8620&52&45\\[-0.05in]
Business&6&10550&85&90\\[-0.05in]
Poetry&6&8016&145&55\\[-0.05in]
Travel&4&11581&112&60\\[-0.05in]
Cricket&4&10044&98&60\\\hline
\multicolumn{5}{|c|}{\textbf{Image Classification}}\\\hline
\textbf{Data set}&\textbf{Classes}&\textbf{Terms}&\textbf{Train}&\textbf{Test}\\\hline
Caltech-101&101&12000&1530&1530\\[-0.05in]
Caltech-tiny&5&12000&75&75\\\hline
\end{tabular}}
\end{table}

Seven thematic TC data sets were considered, in these data sets the goal is to learn a model for thematic categories (e.g., sports news vs. religion news). The considered data sets are the most used ones for the evaluation of TC systems~\citep{sebastiani}. For TC data sets, indexing terms are the words (unigrams). Likewise, seven data sets for AA were used, the goal in these data sets is to learn a model capable of associating documents with authors. Opposed to thematic collections, the goal in AA is to model the writing style of authors, hence, it has been shown that different representations and attributes are necessary for facing this task~\citep{stamatatos}. Accordingly, indexing terms in AA data sets were 3-grams of characters, that is, sequences of 3-characters found in documents, these terms have proved to be the most effective ones in AA~\citep{stamatatos,lowbow,luycx}.
Finally, two data sets for image classification, taken from the CALTECH-101 collection, were used. We considered the  collection under the standard experimental settings (15 images per class for training and 15 images for testing), two subsets of the CALTECH-101 data set were used: a small one with only 5 categories and the whole data set with 102 classes (101 object categories plus background)~\citep{caltech}. Images were represented under the Bag-of-Visual-Words formulation using dense sift descriptors (PHOW features): descriptors extracted from images were clustered using $k-$means, the centers of the clusters are the visual words (indexing terms), images are then represented by accounting the occurrence of visual words, the VLFEAT toolbox was used for processing images~\citep{vedaldi08vlfeat}.

The considered data sets have been partitioned into training and test subsets (the number of documents for each partition and each data set are shown in Table~\ref{tab:datasets}). For some data sets there were predefined categories, while for others we randomly generated them using $70\%$ of documents for training and the rest for testing. All of the preprocessed data sets in Matlab format are publicly available\footnote{\tt{http://ccc.inaoep.mx/~hugojair/TWL/}}.

For each experiment, the training partition was used to learn the TWS, as explained in Section~\ref{sec:method}. The learned TWS is then evaluated in the corresponding test subset. We report two performance measures: accuracy, which is the percentage of correctly classified instances, and $f_1$ measure, which assesses the tradeoff between precision and recall across classes (macro-average $f_1$), recall that $f_1$ was used as fitness function (see Section~\ref{sec:method}).

The genetic program was run for 50 generations using populations of 50 individuals, we would like to point out that in each run of the proposed method we have used default parameters. It is expected that by optimizing parameters and running the genetic program for more generations and larger populations we could obtain even better results. The goal of our study, however, was to show the potential of our method even with default parameters.

\subsection{Evaluation of TWS Learning  via Genetic Programming}
\label{subsec:exp1}
This section reports experimental results on learning TWSs with the genetic program described in Section~\ref{sec:method}. The goal of this experiment is to assess how TWSs learned via GP compare with traditional TWSs. The GP method was run on each of the 16 data sets from Table~\ref{tab:datasets}, since the vocabulary size for some data sets is huge we decided to reduce the number of terms by using term-frequency as criterion. Thus, for each data set we considered the top $2000$ more frequent terms during the search process. In this way, the search process is accelerated at no significant loss of accuracy. In Section~\ref{subsec:vocasize} we analyze the robustness of our method when using the whole vocabulary size for some data sets.

For each data set we performed 5 runs with the GP-based approach, we evaluated the performance of each learned TWS and report the average and standard deviation of performance across the five runs. Tables~\ref{tab:results1}, \ref{tab:results2}, and~\ref{tab:results3} show the performance obtained by TWSs learned for thematic TC, AA and IC data sets, respectively. In the mentioned tables we also show the result obtained by the \emph{best baseline} in each collection. \emph{Best baseline} is the best TWS we found (from the set of TWSs reviewed in related work and the TWSs in Table~\ref{tab:pesados}) for each data set (using the test-set performance). Please note that under these circumstances \emph{best baseline} is in fact, a quite strong baseline for our GP method. Also, we would like to emphasize that no parameter of the GP has been optimized, we used the same default parameters for every execution of the genetic program.
\begin{table}[htb]
\centering
\caption{Classification performance on thematic TC obtained with learned TWSs and the best baseline. }\label{tab:results1}
\footnotesize{
\begin{tabular}{|c|c|c|c|c|c|}
\hline
&\multicolumn{2}{|c|}{\textbf{PG- Avg.}}&\multicolumn{3}{|c|}{\textbf{Best baseline}}\\\hline
Data set&$f_1$& Acc.&$f_1$& Acc.&Baseline\\\hline
Reuters-8&\textbf{90.56}$^+_-$\textbf{1.43}&\textbf{91.35}$^+_-$\textbf{1.99}&86.94&88.63&TF\\
Reuters-10&\textbf{88.21}$^+_-$\textbf{2.69}&91.84$^+_-$1.01&85.24&\textbf{93.25}&TFIDF\\
20-Newsgroup&\textbf{66.23}$^+_-$\textbf{3.84}&\textbf{67.97}$^+_-$\textbf{4.16}&59.21&61.99&TF\\
TDT-2&\textbf{96.95}$^+_-$\textbf{0.41}&\textbf{96.95}$^+_-$\textbf{0.57}&95.20&95.21&TFIDF\\
WebKB&\textbf{88.79}$^+_-$\textbf{1.26}&\textbf{89.12}$^+_-$\textbf{1.30}&87.49&88.62&B\\
Classic-4&\textbf{94.75}$^+_-$\textbf{1.08}&\textbf{95.42}$^+_-$\textbf{0.67}&94.68&94.86&TF\\
CADE-12&\textbf{41.03}$^+_-$\textbf{4.45}&\textbf{53.80}$^+_-$\textbf{4.0}&39.30&41.89&TF\\\hline
\end{tabular}}
\end{table}

From Table~\ref{tab:results1} it can be seen that, regarding the best baseline, different TWSs obtained better performance for different data sets. Hence evidencing the fact that different TWSs are required for different problems. On the other hand, it can be seen that the average performance of TWSs learned with our GP outperformed significantly the best baseline in all but one result (accuracy for Reuters-10 data set). The differences in performance are large, mainly for the $f_1$ measure, which is somewhat expected as this was the measure used as fitness function (recall $f_1$ measure is appropriate to account for the class imbalance across classes); hence showing the competitiveness of our proposed approach for learning effective TWSs for thematic TC tasks.
\begin{table}[htb]
\centering
\caption{Classification performance on AA obtained with learned TWSs and the best baseline. }\label{tab:results2}
\footnotesize{
\begin{tabular}{|c|c|c|c|c|c|}
\hline
&\multicolumn{2}{|c|}{\textbf{PG- Avg.}}&\multicolumn{3}{|c|}{\textbf{Best baseline}}\\\hline
Data set&$f_1$& Acc.&$f_1$& Acc.&Baseline\\\hline
CCA-10&\textbf{70.32}$^+_-$\textbf{2.73}&\textbf{73.72}$^+_-$\textbf{2.14}&65.90&73.15&TF-IG\\
Poetas&\textbf{72.23}$^+_-$\textbf{1.49}&\textbf{72.63}$^+_-$\textbf{1.34}&70.61&71.84&TF-IG\\
Football&76.37$^+_-$9.99&83.76$^+_-$4.27&\textbf{76.45}&\textbf{83.78}&TF-CHI\\
Business&\textbf{78.08}$^+_-$\textbf{4.87}&\textbf{83.58}$^+_-$\textbf{1.57}&73.77&81.49&TF-CHI\\
Poetry&\textbf{70.03}$^+_-$\textbf{7.66}&74.05$^+_-$7.38&59.93&\textbf{76.71}&B\\
Travel&\textbf{73.92}$^+_-$\textbf{10.26}&\textbf{78.45}$^+_-$\textbf{6.72}&71.75&75.32&TF-CHI\\
Cricket&88.10$^+_-$7.12&\textbf{92.06}$^+_-$\textbf{3.29}&\textbf{89.81}&91.89&TF-CHI\\\hline
\end{tabular}}
\end{table}

From Table~\ref{tab:results2} it can be seen that for AA data sets the best baseline performs similarly as the proposed approach. In terms of $f_1$ measure, our method outperforms the best baseline in 5 out of 7 data sets, while in accuracy our method beats the best baseline in 4 out of 7 data sets. Therefore, our method still obtains comparable (slightly better) performance to the best baselines, which for AA tasks were much more competitive than in thematic TC problems. One should note that for PG we are reporting the average performance across 5 runs, among the 5 runs we found TWSs that consistently outperformed the best baseline.
\begin{table}[htb]
\centering
\caption{Classification performance on IC obtained with learned TWSs and the best baseline. }\label{tab:results3}
\footnotesize{
\begin{tabular}{|c|c|c|c|c|c|}
\hline
&\multicolumn{2}{|c|}{\textbf{PG- Avg.}}&\multicolumn{3}{|c|}{\textbf{Best baseline}}\\\hline
Data set&$f_1$& Acc.&$f_1$& Acc.&Baseline\\\hline
Caltech-101&\textbf{61.91}$^+_-$\textbf{1.41}&\textbf{64.02}$^+_-$\textbf{1.42}&58.43&60.28&B\\
Caltech-tiny&\textbf{89.70}$^+_-$\textbf{2.44}&\textbf{91.11}$^+_-$\textbf{2.36}&85.65&86.67&TF\\\hline
\end{tabular}}
\end{table}

It is quite interesting that, comparing the best baselines from Tables~\ref{tab:results1} and~\ref{tab:results2}, for AA tasks supervised TWSs obtained the best results (in particular TF-CHI in 4 out of 7 data sets), whereas for thematic TC unsupervised TWSs performed better. Again, these results show that different TWSs are required for different data sets and different types of problems. In fact, our results confirm the fact that AA and thematic TC tasks are quite different, and, more importantly, our study provides evidence on the suitability of supervised TWSs for AA; to the best of our knowledge, supervised TWSs have not been used in AA problems.

Table~\ref{tab:results3} shows the results obtained for the image categorization data sets. Again, the proposed method obtained TWSs that outperformed the best baselines. This result is quite interesting because we are showing that the TWS plays a key role in the classification of images under the BOVWs approach. In computer vision most of the efforts so far have been devoted to the development of novel/better low-level image-descriptors, using a BOW with predefined TWS.
Therefore, our results pave the way for research on learning TWSs for image categorization and other tasks that rely in the BOW representation (e.g. speech recognition and video classification).

\begin{figure}[htb!]
    \centering
    \includegraphics[scale=0.6]{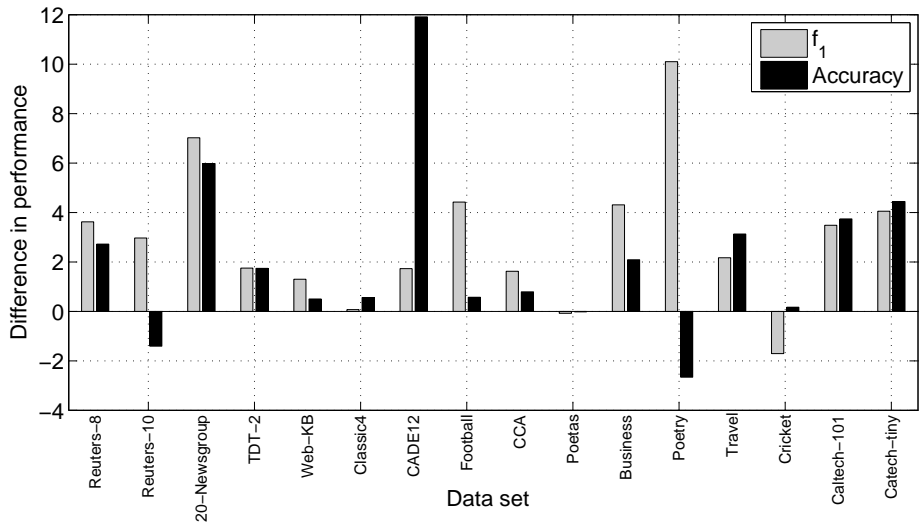}
    \caption{Difference in performance between learned TWSs and best baseline per each data set, values above zero indicate better performance obtained by the TWSs. }
  \label{fig:diffperf}
\end{figure}

Figure~\ref{fig:diffperf} and Table~\ref{tab:results4} complement the results presented so far.  Figure~\ref{fig:diffperf} indicates the difference in performance between the (average of) learned TWSs and the best baseline for each of the considered data sets. We can clearly appreciate from this figure the magnitude of improvement offered by the learned TWSs, which in some cases is too large.

Table~\ref{tab:results4}, on the other hand, shows a more fair comparison between our method and the reference TWSs: it shows the average performance obtained by reference schemes and the average performance of our method for thematic TC, AA and IC data sets. It is clear from this table that in average our method performs consistently better than any of the reference methods in terms of both accuracy and $f_1$ measure for the three types of tasks. Thus, from the results of this table and those from Tables~\ref{tab:results1}, \ref{tab:results2}, and~\ref{tab:results3}, it is evident that standard TWSs are competitive, but one can take advantage of them only when the right TWS is selected for each data set. Also, the performance of TWSs learned with our approach are a better option than standard TWSs, as in average we were able to obtain much better representations.
\begin{table}[htb]
\centering
\caption{Average performance on thematic TC obtained with learned TWSs and the baselines. }\label{tab:results4}
\footnotesize{
\begin{tabular}{|c|c|c|c|c|c|c|}
\hline
&\multicolumn{2}{|c|}{\textbf{Thematic TC}}&\multicolumn{2}{|c|}{\textbf{AA}}&\multicolumn{2}{|c|}{\textbf{IC}}\\\hline
TWS&$f_1$& Acc.&$f_1$& Acc.&$f_1$& Acc.\\\hline
TF&76.60&79.53&62.17&72.43&68.86&71.54\\
B&77.42&79.73&66.07&76.76&71.22&72.78\\
TFIDF&61.69&76.17&40.88&55.26&62.27&67.56\\
TF-CHI&71.56&75.63&68.75&73.69&65.38&67.45\\
TF-IG&64.22&69.00&68.96&74.91&66.02&67.93\\\hline
PG-worst&77.81&81.19&66.47&74.84&74.30&75.67\\
PG-Avg.&\textbf{81.01}&\textbf{83.63}&\textbf{75.58}&\textbf{79.75}&\textbf{75.81}&\textbf{77.07}\\
PG-best&82.88&85.81&81.37&83.98&76.97&78.18\\\hline
\end{tabular}}
\end{table}

Summarizing the results from this section, we can conclude that:
\begin{itemize}
\item The proposed GP obtained TWSs that outperformed the best baselines in the three types of tasks: thematic TC, AA and IC. Evidencing the generality of our proposal across different data types and modalities. Larger improvements were observed for thematic TC and IC data sets. In average, learned TWSs outperformed standard ones in the three types of tasks.

\item Our results confirm our hypothesis that different TWSs are required for facing different tasks, and within a same task (e.g., AA) a different TWS may be required for a different data set. Hence, motivating further research on how to select TWS for a particular TC problem.

\item We show evidence that the proposed TWS learning approach is a promising solution for enhancing the classification performance in other tasks than TC, e.g., IC.

\item Our results show that for AA supervised TWS seem to be more appropriate, whereas unsupervised TWS performed better on thematic TC and IC. This is a quite interesting result that may have an impact in non-thematic TC and supervised term-weighting learning.
\end{itemize}

\subsection{Varying vocabulary size}
\label{subsec:vocasize}

For the experiments from Section~\ref{subsec:exp1} each TWS was learned by using only the top 2000 most frequent terms during the search process. This reduction in the vocabulary allowed us to speed up the search process significantly, however, it is worth asking ourselves what the performance of the TWSs would be when using an increasing number of terms. We aim to answer such question in this section.

\begin{figure}[htb!]
    \centering
            \includegraphics[width=6.5cm,height=4cm]{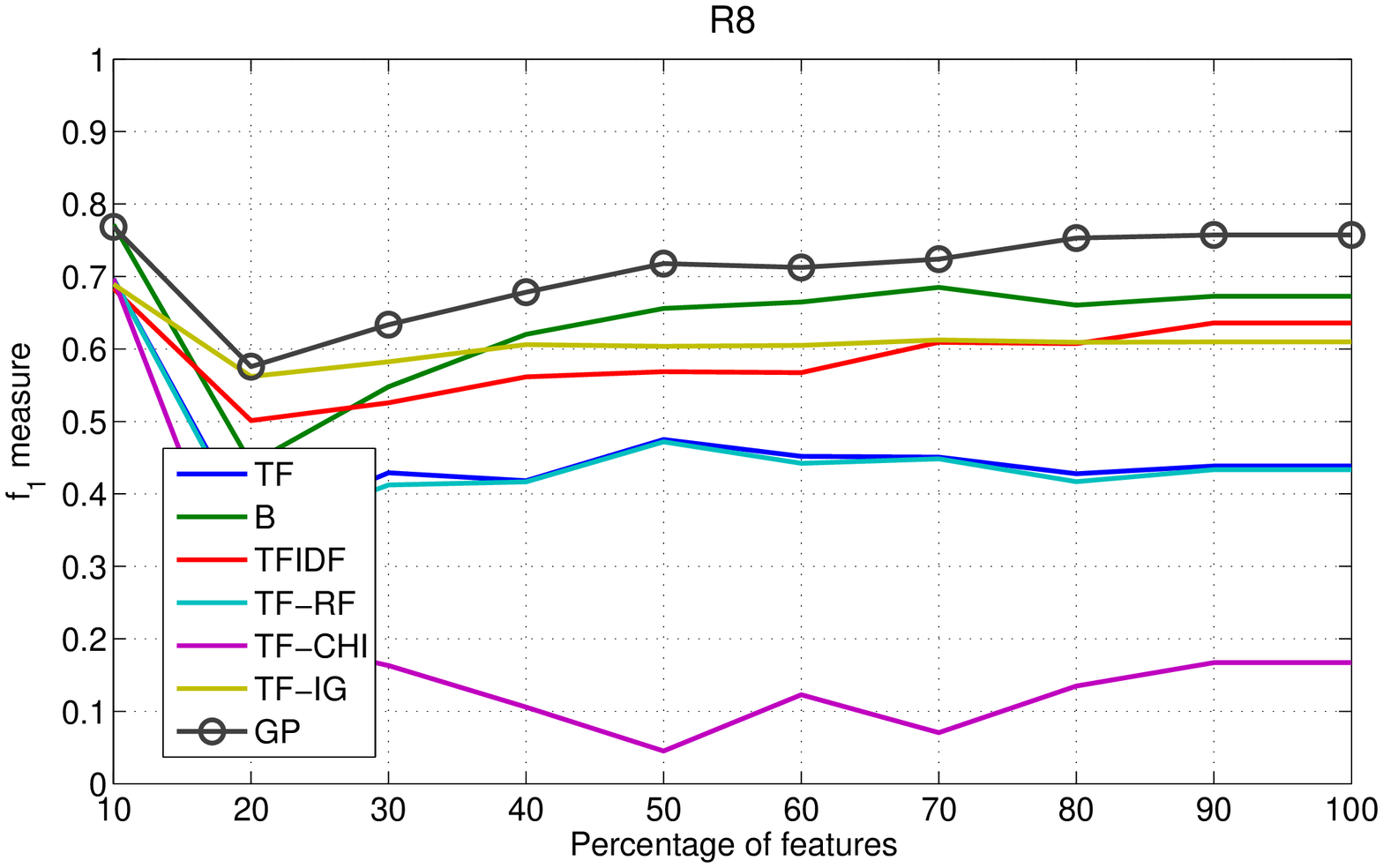}
    \includegraphics[width=6.5cm,height=4cm]{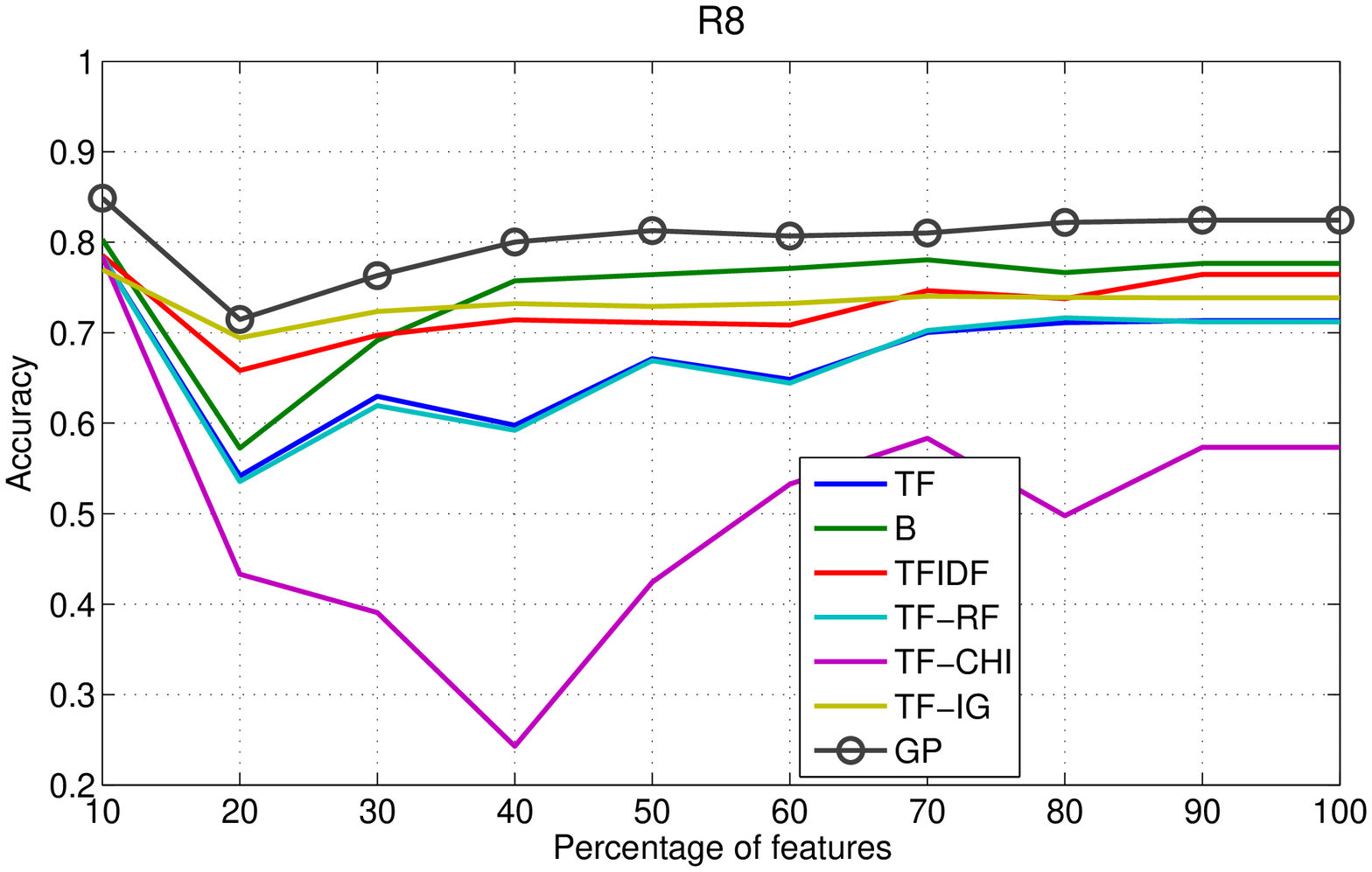}
    \caption{Classification performance on the Reuters-8 data set for the TWS: $\sqrt{\mathbf{W}_{5}}-\frac{\log{\sqrt{\mathbf{W}_{19}}}}{\mathbf{W}_{21}}$ 
    when increasing the number of considered terms.  The left plot shows results in terms of $f_1$ measure while the right plot shows accuracy performance. }
  \label{fig:varyingsizes}
\end{figure}

\begin{figure}[htb!]
    \centering
        \includegraphics[width=6.5cm,height=4cm]{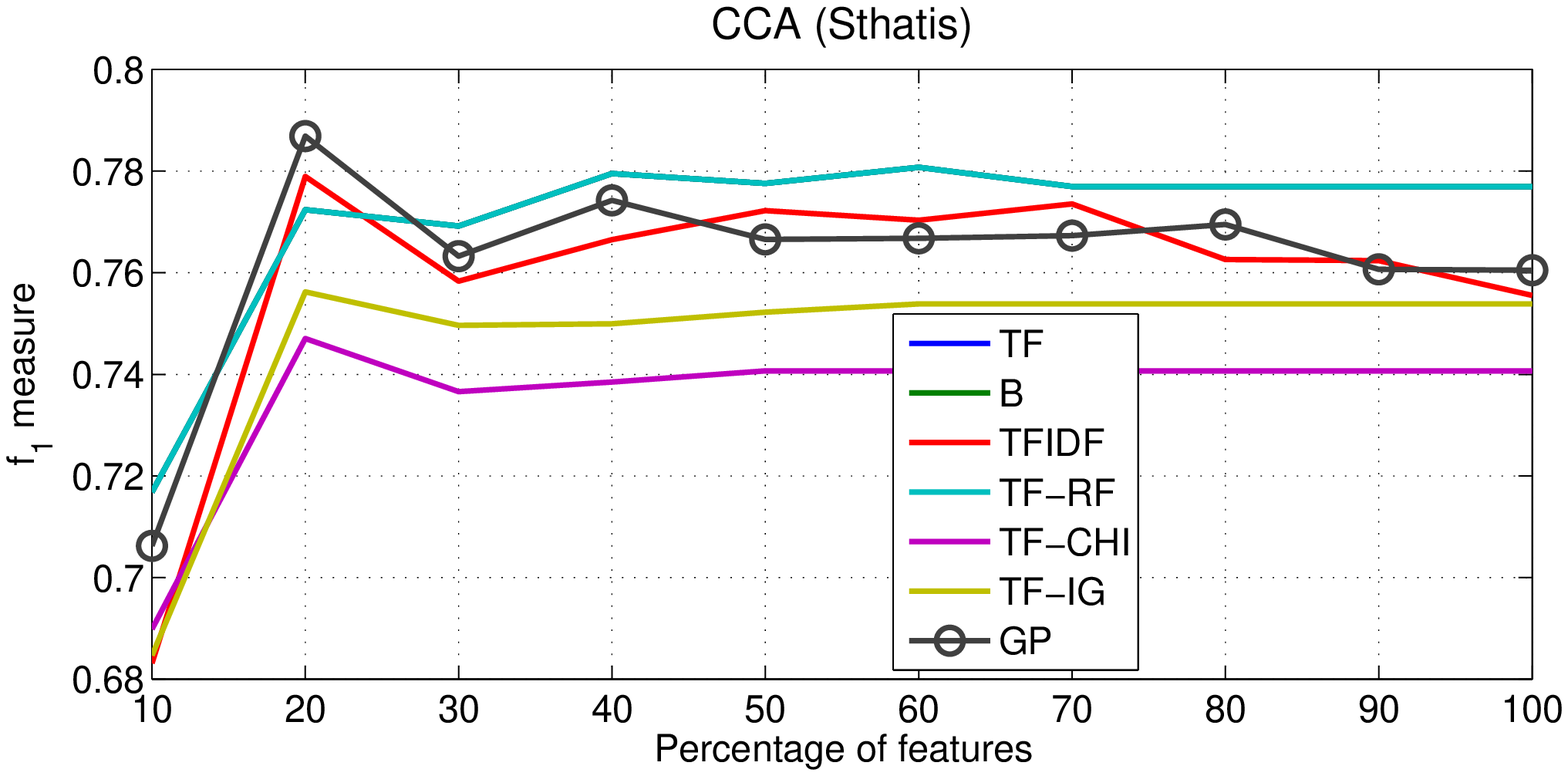}
  \includegraphics[width=6.5cm,height=4cm]{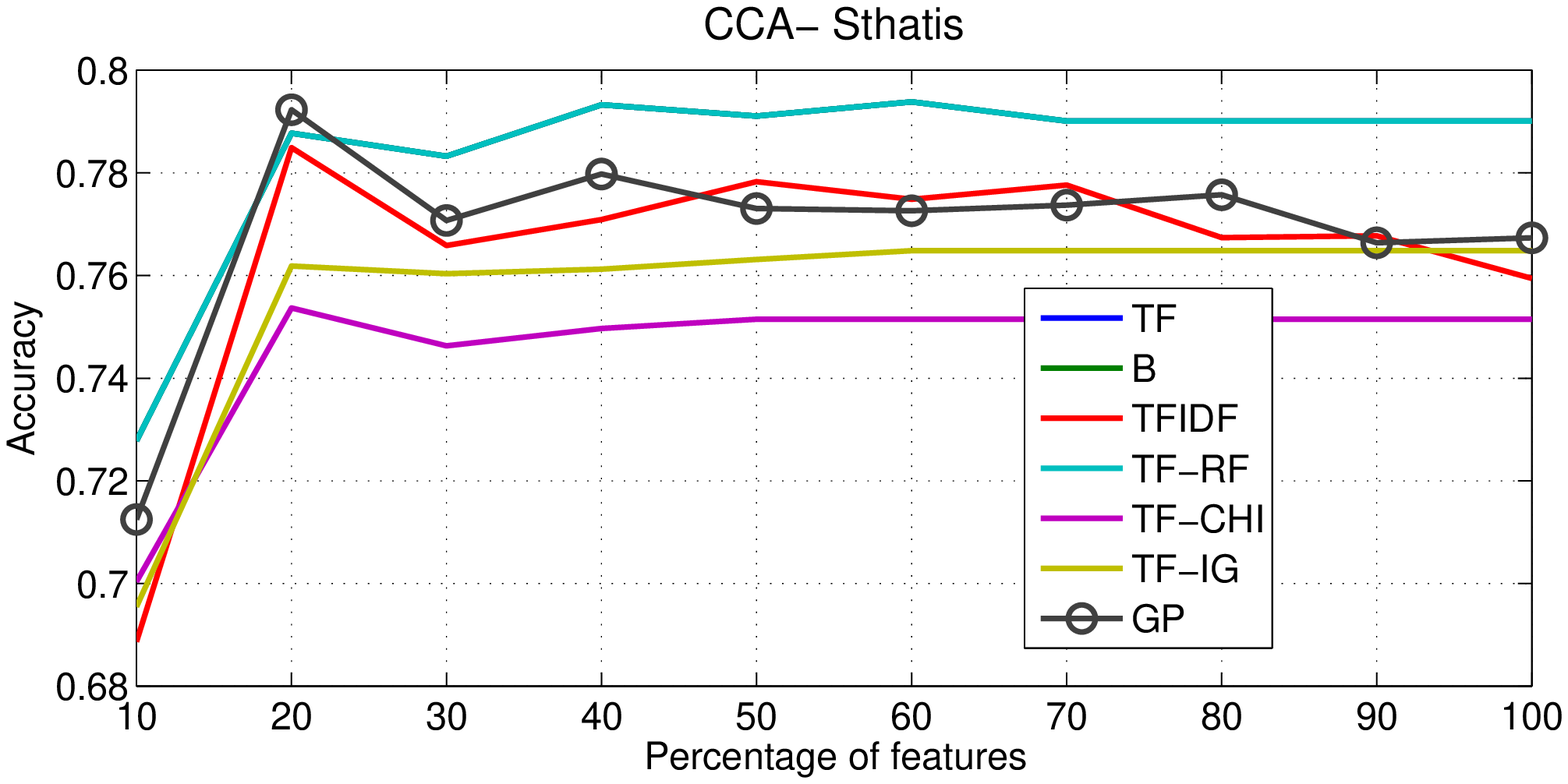}
    \caption{Classification performance on the CCA data set of the TWS: $\mathbf{W}_{4}-(\mathbf{W}_{22}+\mathbf{W}_{5})$ 
    when increasing the number of considered terms.  }
  \label{fig:varyingsizes2}
\end{figure}

\begin{figure}[htb!]
    \centering
   \includegraphics[width=6.5cm,height=4cm]{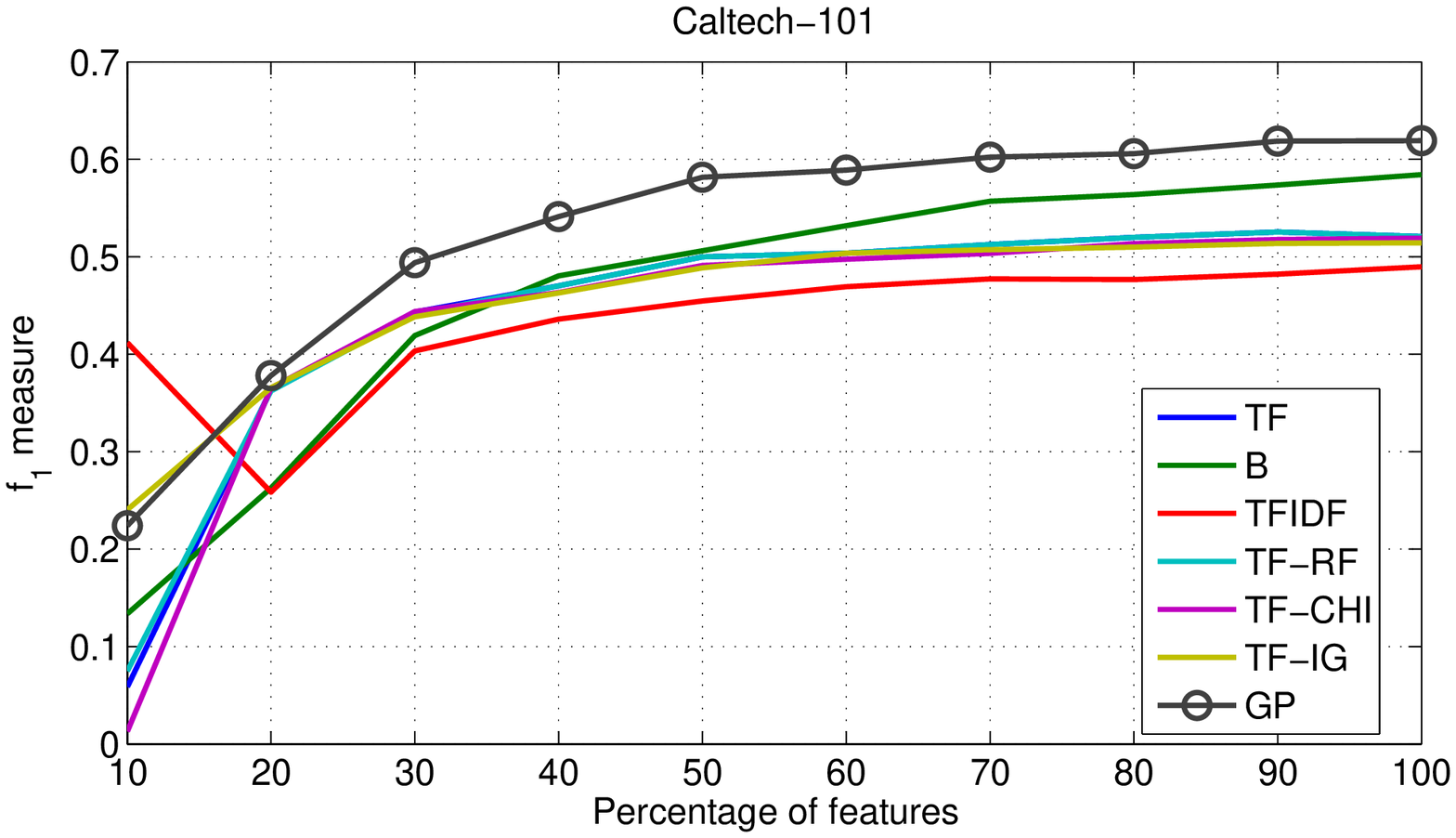}
   \includegraphics[width=6.5cm,height=4cm]{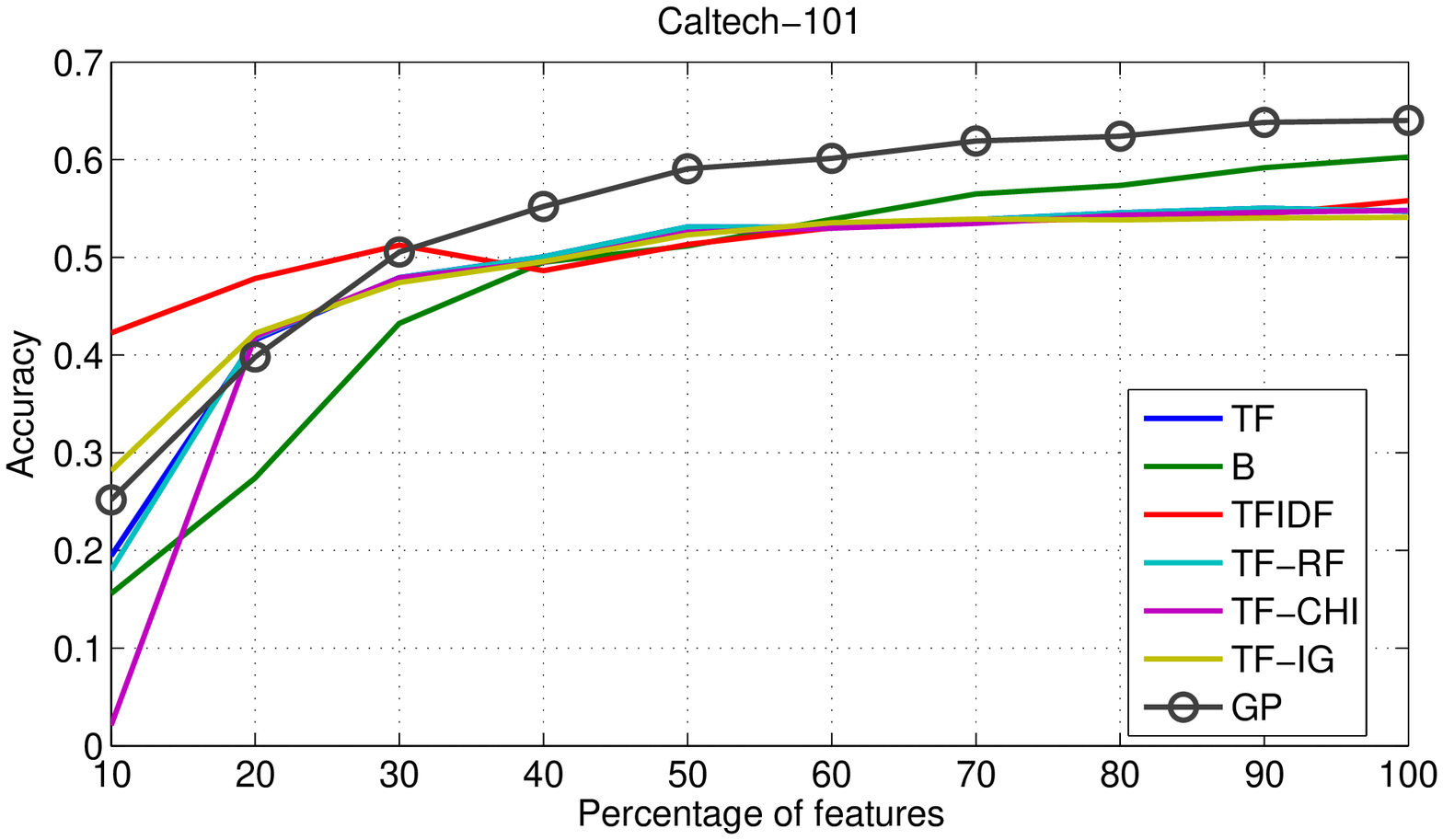}
    \caption{Classification performance on the Caltech-101 data set of the TWS: $\sqrt{\sqrt{\mathbf{W}_{17}-\sqrt{\sqrt{\mathbf{W}_{22}}}}}$ 
    when increasing the number of considered terms.  }
  \label{fig:varyingsizes4}
\end{figure}

For this experiment we considered three data sets, one from each type of task: thematic TC, AA, and IC. The considered data sets were the Reuters-8 (R8) for thematic TC, the CCA benchmark for AA, and Caltech-101 for IC. These data sets are the representative ones from each task: Reuters-8 is among the most used TC data sets, CCA has been widely used for AA as well, and Caltech-101 is \emph{the benchmark} in image categorization For each of the considered data sets we use a specific TWS learned using the top-2000 most frequent terms (see Section~\ref{subsec:exp1}), and evaluate the performance of such TWSs when increasing the vocabulary size: terms were sorted in ascending order of their frequency.  Figures~\ref{fig:varyingsizes},~\ref{fig:varyingsizes2}, and~\ref{fig:varyingsizes4} show the results of this experiment in terms of $f_1$ measure and accuracy (the selected TWS is shown in the caption of each image).

Different performance behavior can be observed in the different data sets. Regarding Figure~\ref{fig:varyingsizes}, which shows the performance for a thematic TC data set, it can be seen that the TWS learned by our method outperformed all other TWSs for any data set size. Hence confirming the suitability of the proposed method for thematic TC.

Figure~\ref{fig:varyingsizes2}, on the other hand, behaves differently: the proposed method outperforms all the other TWSs only for a single data set size (when $20\%$ of the terms were used). In general, our method consistently outperformed TF-CHI and TF-IG TWSs,  and performs similarly as TF-IDF, but it was outperformed by the TF-RF TWS. This result can be due to the fact that for this AA data set, the genetic program learned a TWS that was suitable only for the vocabulary size that was used during the optimization. Although, interesting, this result is not that surprising: in fact, it is well known in AA that the number of terms considered in the vocabulary plays a key role on the performance of AA systems. AA studies suggest using a small amount of the most-frequent terms when approaching an AA problem~\citep{stamatatos,lowbow,luycx}. Results from Figure~\ref{fig:varyingsizes2} corroborate the latter and seem to indicate that when approaching an AA problem, one should first determine an appropriate vocabulary size and then apply our method. One should note, however, that our method outperforms the other TWSs for the data set size that was used during the optimization, and this is, in fact, the highest performance that can be obtained with any other TWS and data set size combination.

Finally, Figure~\ref{fig:varyingsizes4} reports the performance of TWSs for the Caltech-101 data set under different data set sizes. In this case, the learned TWS outperforms all other TWSs when using more than $20\%$ and $30\%$ in terms of $f_1$ measure and accuracy, respectively. The improvement is consistent and monotonically increases as more terms are considered. Hence showing the robustness of the learned TWS when increasing the vocabulary size for IC tasks. Among the other TWSs, TFIDF obtains competitive performance when using a small vocabulary, this could be due to the fact that when considering a small number of frequent terms the IDF component is important for weighting the contribution of each of the terms.

Summarizing the results from this section we can conclude the following:
\begin{itemize}
\item TWSs learned with our method are robust to variations in the vocabulary size for thematic TC and IC tasks. This result suggests, we can learn TWSs using a small number of terms (making more efficient the search process) and evaluating the learned TWSs with larger vocabularies.

\item Learned TWSs outperform standard TWSs in thematic TC and IC tasks when varying the vocabulary size.

\item For AA, TWSs learned with our proposed approach seem to be more dependent on the number of terms used during training. Hence, when facing this type of problems it is a better option to fix the number of terms beforehand and then running our method.

\end{itemize}

\subsection{Generalization of the learned term-weights}
In this section we evaluate the inter-data set generalization capabilities of the learned TWSs. Although results presented so far show the generality of our method across three types of tasks, we have reported results obtained with TWSs that were learned for each specific data set. It remains unclear whether the TWSs learned for a collection can perform similarly in other collections, we aim to answer to this question in this section.

To assess the inter-data set generalization of TWSs learned with our method we performed an experiment in which we considered for each data set a single TWS and evaluated its performance across all the 16 considered data sets. The considered TWSs are shown in Table~\ref{tab:selpesados}, we named the variables with meaningful acronyms for clarity but also show the mathematical expression using variables as defined in Table~\ref{tabla:params}.
\begin{table}[htb]
\centering
\caption{Considered TWSs for the inter-data set generalization experiment for each data set. In column 2 each TWS is shown as a prefix expression, the names of the variables are self-explanatory. Column 3 shows the mathematical expression of each TWS using the terminal set from Table~\ref{tabla:params}.}\label{tab:selpesados}
\tiny{
\begin{tabular}{|p{0.5cm}|p{1.3cm}|p{6.5cm}|p{4cm}|}
\hline
\multicolumn{4}{|c|}{\textbf{Text categorization}}\\\hline
\textbf{ID}&\textbf{Data set}&\textbf{Learned TWS}&\textbf{Formula}\\\hline
1&Reuters-8&-(sqrt(TFIDF),div(log2(sqrt(ProbR)),RF))&$\sqrt{\mathbf{W}_{5}}-\frac{\log{\sqrt{\mathbf{W}_{19}}}}{\mathbf{W}_{21}}$\\
2&Reuters-10&sqrt(div(pow2(sqrt(TFIDF)),div(pow2(TF-RF),pow2(TF-RF))))&$\sqrt{\frac{(\sqrt{\mathbf{W}_{5}})^2}{\frac{\mathbf{W}_{22}^2}{\mathbf{W}_{22}^2}}}$\\
3&20-Newsg.&sqrt(sqrt(div(TF,GLOBTF)))&$\sqrt{\sqrt{\frac{\mathbf{W}_{6}}{\mathbf{W}_{7}}}}$\\
4&TDT-2&sqrt($\times$(sqrt(sqrt(TFIDF)), sqrt($\times$(sqrt(TFIDF),IG))))&$\sqrt{\sqrt{\sqrt{\mathbf{W}_{5}}} \times \sqrt{\sqrt{\mathbf{W}_{5}}\times \mathbf{W}_{4}}}$\\
5&WebKB&div(TF-RF,+(+(+(RF,TF-RF),FMEAS),FMEAS))&$\frac{\mathbf{W}_{22}}{(((\mathbf{W}_{21}+\mathbf{W}_{22})+\mathbf{W}_{16})+\mathbf{W}_{16})}$\\
6&Classic-4&$\times$(ProbR,TFIDF)&$\mathbf{W}_{5}\times\mathbf{W}_{19}$\\
7&CADE-12&div(TF,sqrt(log2(ACCU)))&$\frac{\mathbf{W}_{6}}{\sqrt{\log{\mathbf{W}_{12}}}}$\\\hline
\multicolumn{4}{|c|}{\textbf{Authorship attribution}}\\\hline
\textbf{ID}&\textbf{Data set}&\textbf{Learned TWS}&\textbf{Formula}\\\hline
8&CCA-10&-(IG,plus(TF-RF,TFIDF))&$\mathbf{W}_{4}-(\mathbf{W}_{22}+\mathbf{W}_{5})$\\
9&Poetas&-(-(RF,TF-RF),TF-IDF)&$(\mathbf{W}_{21}-\mathbf{W}_{22})-\mathbf{W}_{5}$\\
10&Football&div(TF-RF,pow2(ODDSR))&$\frac{\mathbf{W}_{22}}{\mathbf{W}_{17}^2}$\\
11&Business&minus(TF-RF,PROBR)&$\mathbf{W}_{22}-\mathbf{W}_{19}$\\
12&Poetry&div(TF,log2(div(TF,log(TF-RF))))&$\frac{\mathbf{W}_{6}}{\log{\frac{\mathbf{W}_{6}}{\log{\mathbf{W}_{22}}}}}$\\
13&Travel&+(-(TF-RF,-(-(TF-RF,-(TF-RF,POWER)),POWER)),TF)&$ (\mathbf{W}_{22}-((\mathbf{W}_{22}-(\mathbf{W}_{22}-\mathbf{W}_{18}))-\mathbf{W}_{18}))+\mathbf{W}_{6}$\\
14&Cricket&$\times$(IG,TF-RF)&$\mathbf{W}_{4}\times \mathbf{W}_{22}$\\\hline
\multicolumn{4}{|c|}{\textbf{Image Classification}}\\\hline
\textbf{ID}&\textbf{Data set}&\textbf{Learned TWS}&\textbf{Formula}\\\hline
15&Caltech-101&sqrt(sqrt(-(ODDSR,sqrt(sqrt(TF-RF)))))&$\sqrt{\sqrt{\mathbf{W}_{17}-\sqrt{\sqrt{\mathbf{W}_{22}}}}}$\\
16&Caltech-tiny&sqrt(-(TF-RF,ACBAL))&$\sqrt{\mathbf{W}_{22} - \mathbf{W}_{13}}$\\\hline
\end{tabular}}
\end{table}

Before presenting the results of the experiments it is worth analyzing the type of solutions (TWSs) learned with the proposed approach. First of all, it can be seen that the learned TWSs are not too complex: the depth of the trees is small and solutions have few terminals as components. This is a positive result because it allows us to better analyze the solutions and, more importantly, it is an indirect indicator of the absence of the over-fitting phenomenon. Secondly, as in other applications of genetic programming, it is unavoidable to have unnecessary terms in the solutions, for instance, the subtree: \emph{div(pow2(TF-RF),pow2(TF-RF)))}, (from TWS 2) is unnecessary because it reduces to a constant matrix; the same happens with the term \emph{pow2(sqrt(TFIDF))}. Nevertheless, it is important to emphasize that this type of terms do not harm the performance of learned TWSs, and there are not too many of these type of subtrees.
On the other hand, it is interesting that all of the learned TWSs incorporate supervised information. The most used TR weight is RF, likewise the most used TDR is TFIDF. Also it is interesting that simple operations over standard TWSs, TR and TDR weights results in significant performance improvements. For instance, compare the performance of TF-RF and the learned weight for Caltech-101 in Figure~\ref{fig:varyingsizes4}. By simply subtracting an odds-ratio from the TF-RF TWS and applying scaling operations, the resultant TWS outperforms significantly TF-RF.

\begin{figure}[htb!]
    \centering
        \includegraphics[scale=0.6]{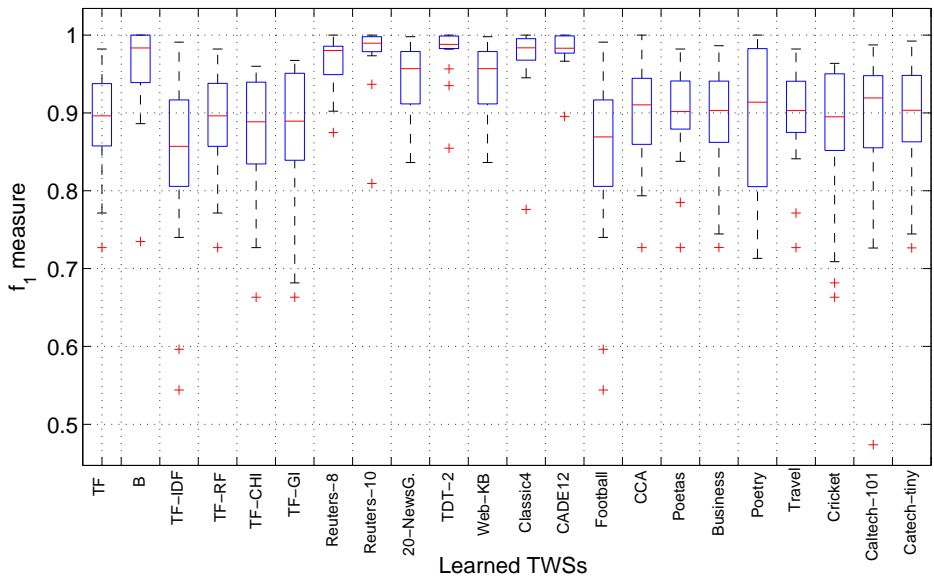}
    \caption{Heatmap that shows the performance of TWSs (rows 7-22) from Table~\ref{tab:selpesados} in the 16 data sets (x axis) considered in the study. For completion, we also show the performance of standard TWSs (rows 1-6).}
  \label{fig:crossd1}
\end{figure}

The 16 TWSs shown in Table~\ref{tab:selpesados} were evaluated in the 16 data sets in order to determine the inter-data set generality of the learned TWSs. Figure~\ref{fig:crossd1} shows the results of this experiment. We show the results with boxplots, where  
each boxplot indicates the normalized\footnote{Before generating the boxplots we normalized the performance on a per-data set basis: for each data set, the performance of the 16 TWSs was normalized to the range $[0,1]$, in this way, the variation in $f_1$-values across data sets is eliminated,  i.e, all $f_1$ values are in the same scale and are comparable to each other.} performance of each TWSs across the 16 data sets, for completion we also show the performance of the reference TWSs on the 16 data sets.  

It can be seen from Figure~\ref{fig:crossd1} that the generalization performance of learned TWSs is mixed.  
On the one hand, it is clear that TWSs learned for thematic TC (boxplots 7-13) achieve the highest generalization performance. Clearly, the generalization performance of these TWSs is higher than that of traditional TWSs (boxplots 1-6). It is interesting that TWSs learned for a particular data set/problem/modality perform well across different data sets/problems/modalities. In particular, TWSs learned for \emph{Reuters-10} and \emph{TDT-2} obtained the highest performance and the lowest variance among all of the TWSs. On the other hand, TWSs learned for AA and IC tasks obtained lower generalization performance: the worst in terms of variance is the TWS learned for the \emph{Poetry} data set, while the worst average performance was obtained by the TWS learned for the \emph{Football} data set. TWSs learned for IC are competitive (in generalization performance) with traditional TWSs. Because of the nature of the tasks, the generalization performance of TWSs learned from TC is better than that of TWSs learned for AA and IC.
One should note that these results confirm our findings from previous sections: (i) the proposed approach is very effective mainly for thematic TC and IC tasks; and, (ii) AA data sets are difficult to model with TWSs.  

\begin{figure}[!htb]
    \centering
            \includegraphics[scale=0.45]{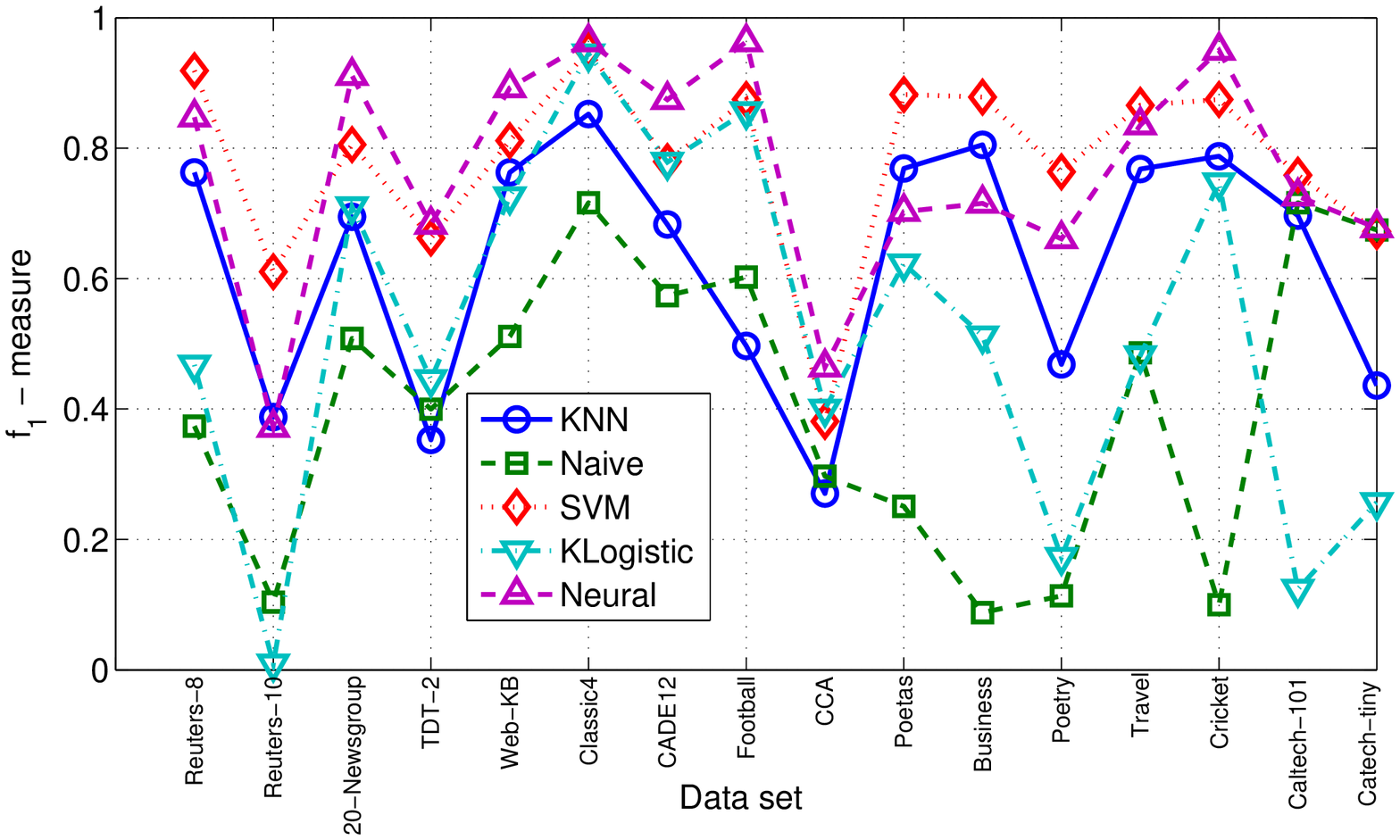}
        \includegraphics[scale=0.45]{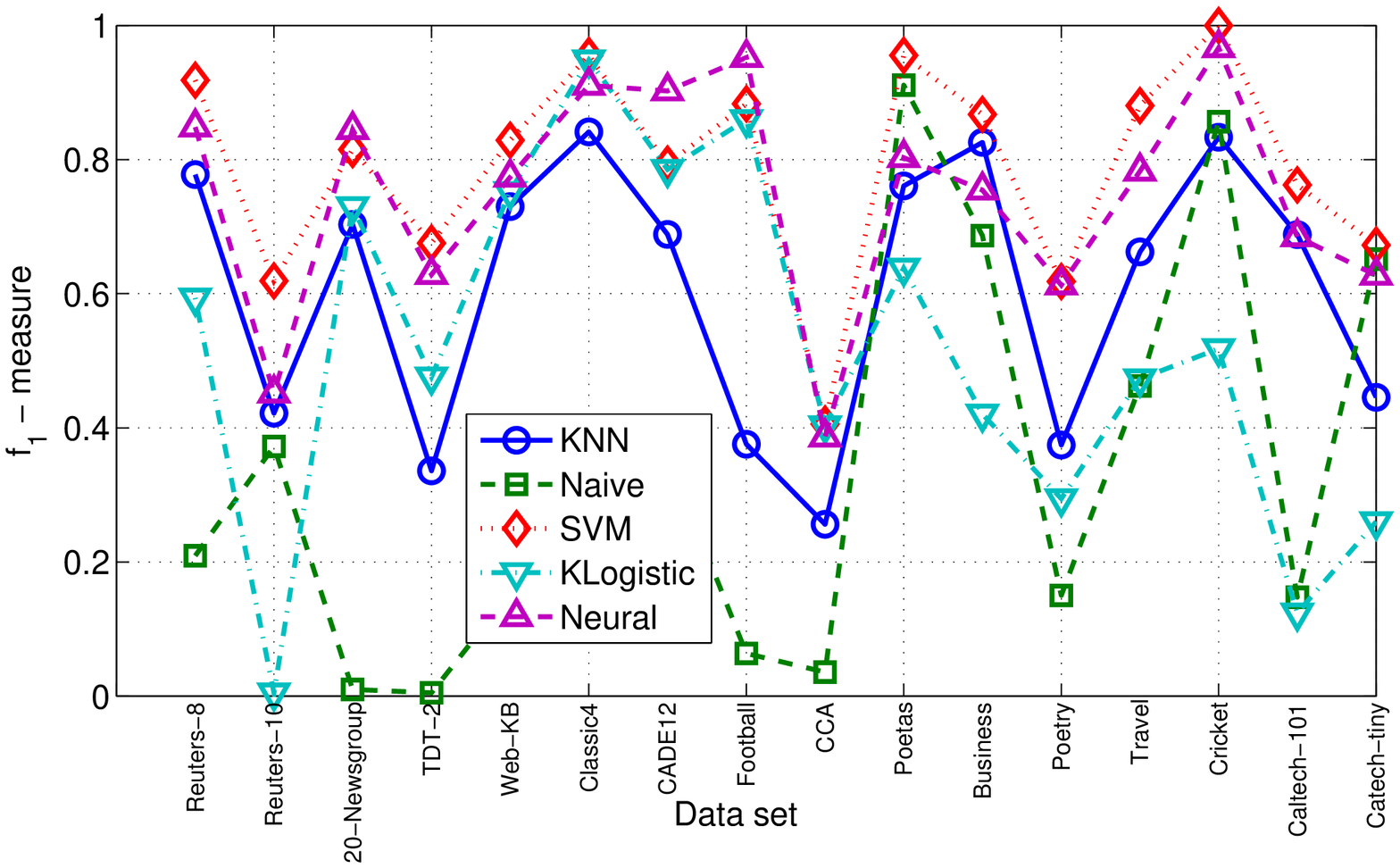}
    \caption{Classification performance of selected TWSs across different classifiers, $f_1$ measure is reported. The plot at the top is a TWS learned for Caltech-tiny, while the bottom plot shows the performance for a TWS learned for Caltech-101.}
  \label{fig:geneclass}
\end{figure}

Finally, we evaluate the generality of learned TWSs across different classifiers. The goal of this experiment is to assess the extend to which the learned TWSs are tailored for the classifier they were learn for. For this experiment, we selected two TWSs corresponding to Caltech-tiny and Caltech-101 (15 and 16 in Table~\ref{tab:selpesados}) and evaluated their performance of different classifiers across the 16 data sets. Figure~\ref{fig:geneclass} shows the results of this experiment.

It can be seen from Figure~\ref{fig:geneclass} that the considered TWSs behaved quite differently depending on the classifier. On the one hand, the classification performance when using na\"ive Bayes (\emph{Naive}), kernel-logistic regression (KLogistic), and $1-$nearest neighbors (\emph{KNN}) classifiers degraded significantly. On the other hand, the performance SVM and the neural network (NN) was very similar. These results show that TWSs are somewhat robust across classifiers of similar nature as SVM and NN are very similar classifiers: 
both are linear models in the parameters. The other classifiers are quite different to the reference SVM and, therefore, the performance is poor\footnote{One should note that among the three worse classifiers, KNN, Naive and KLogistic, the latter obtained better performance than the former two, this is due to the fact that KLogistic is closer, in nature, to an SVM. }. It is interesting that in some cases the NN classifier outperformed the SVM, although in average the SVM performed better. This is a somewhat expected result as the performance of the SVM was used as fitness function.

According to the experimental results from this section we can draw the following conclusions:
\begin{itemize}
\item TWSs learned with the proposed approach are not too complex despite their effectiveness. Most of the learned TWSs included a supervised component, evidencing the importance of taking advantage of labeled documents.

\item TWSs offer acceptable inter-data set generalization performance, in particular, TWSs learned for TC generalize pretty well across data sets.

\item We showed evidence that TWSs learned for a modality (e.g., text / images)     can be very competitive when evaluated on other modality.

\item TWSs are somewhat robust to the classifier choice. It is preferable to use the classifier used to estimate the fitness function, although classifiers of similar nature perform similarly.
\end{itemize}

\section{Conclusions}
\label{sec:conclusions}
We have described a novel approach to term-weighting scheme (TWS) learning in text classification (TC). TWSs specify the way in which documents are represented under a vector space model. We proposed a  genetic programming solution in which standard TWSs, term-document, and term relevance weights are combined to give rise to effective TWSs. We reported experimental results in 16 well-known data sets comprising thematic TC, authorship attribution and image classification tasks. The performance of the proposed method is evaluated under different scenarios. Experimental results show that the proposed approach learns very effective TWSs that outperform standard TWSs.
The main findings of this work can be summarized as follows:
\begin{itemize}
\item TWSs learned with the proposed approach outperformed significantly to standard TWSs and those proposed in related work.
\item Defining the appropriate TWS is crucial for image classification tasks, an ignored  issue in the field of computer vision.
\item In authorship attribution, supervised TWSs are beneficial, in comparison with standard TWSs.
\item The performance of learned TWSs do not degrades when varying the vocabulary size for thematic TC and IC. For authorship attribution a near-optimal vocabulary size should be selected before applying our method.
\item TWSs learned for a particular data set or modality can be applied to other data sets or modalities without degrading the classification performance. This generalization capability is mainly present in TWSs learned for thematic TC and IC.
\item Learned TWSs are easy to analyze/interpret and do not seem to overfit the training data.
\end{itemize}

Future work directions include studying the suitability of the proposed approach to learn weighting schemes for cross domain TC. Also we would like to perform an in deep study on the usefulness of the proposed GP for computer vision tasks relying in the Bag-of-Visual-Words formulation.
\bibliographystyle{elsarticle-harv}

\end{document}